\documentclass[10pt,twocolumn,letterpaper]{article}

\usepackage{cvpr}              
\definecolor{cvprblue}{rgb}{0.21,0.49,0.74}
\usepackage[pagebackref,breaklinks,colorlinks,allcolors=cvprblue]{hyperref}

\def\paperID{***} 
\def\confName{CVPR}
\def\confYear{2026}

\title{SemVideo: Reconstructs What You Watch from Brain Activity via Hierarchical Semantic Guidance}

\author{
\textbf{Minghan Yang}$^{1}$,
\textbf{Lan Yang}$^{1}$\thanks{Corresponding author} ,
\textbf{Ke Li}$^1$,
\textbf{Honggang Zhang}$^1$,
\textbf{Kaiyue Pang}$^2$,
\textbf{Yizhe Song}$^2$,
\\
$^1$Beijing University of Posts and Telecommunications \\
$^2$ University of Surrey \\
\texttt{\{yangminghan, ylan, like1990, zhhg\}@bupt.edu.cn} \\
\texttt{hatkpang@gmail.com, y.song@surrey.ac.uk}
}
\begin{document}
\maketitle
\newcommand{\Semprompt}{\textit{SemMiner}}
\newcommand{\Semvideo}{\textit{SemVideo}}
\begin{abstract}
Reconstructing dynamic visual experiences from brain activity provides an avenue for exploring the neural mechanisms of visual perception. While progress in fMRI-based image reconstruction has been notable, extending this success to video reconstruction remains a challenge. Current fMRI-to-video reconstruction approaches encounter two major shortcomings: (i) inconsistent visual representations of salient objects across frames, leading to appearance mismatches; (ii) poor temporal coherence, resulting in motion misalignment or abrupt frame transitions.
To address these limitations, we introduce \textbf{\Semvideo}, a fMRI-to-video reconstruction framework guided by hierarchical semantic information. 
At the core of \Semvideo~is \textbf{\Semprompt}, a hierarchical guidance module that constructs three levels of semantic cues from the video stimulus: static anchor descriptions, motion-oriented narratives, and holistic summaries. Leveraging this semantic guidance, \Semvideo~comprises three key components: a Semantic Alignment Decoder that aligns fMRI signals with CLIP-style embeddings derived from \Semprompt, a Motion Adaptation Decoder that reconstructs dynamic motion patterns using a novel tripartite attention fusion architecture, and a Conditional Video Render that leverages hierarchical semantic guidance for video reconstruction. Experiments on the CC2017 and HCP datasets demonstrate that \Semvideo~achieves superior performance in semantic alignment and temporal consistency, setting a new state-of-the-art in fMRI-to-video reconstruction.  The code is publicly available at \url{https://github.com/yang-minghan/SemVideo}.
\end{abstract}    
\section{Introduction}
\label{sec:intro}
\begin{figure}[htbp]
    \includegraphics[width=0.98\linewidth, trim=0 0 0 0, clip]{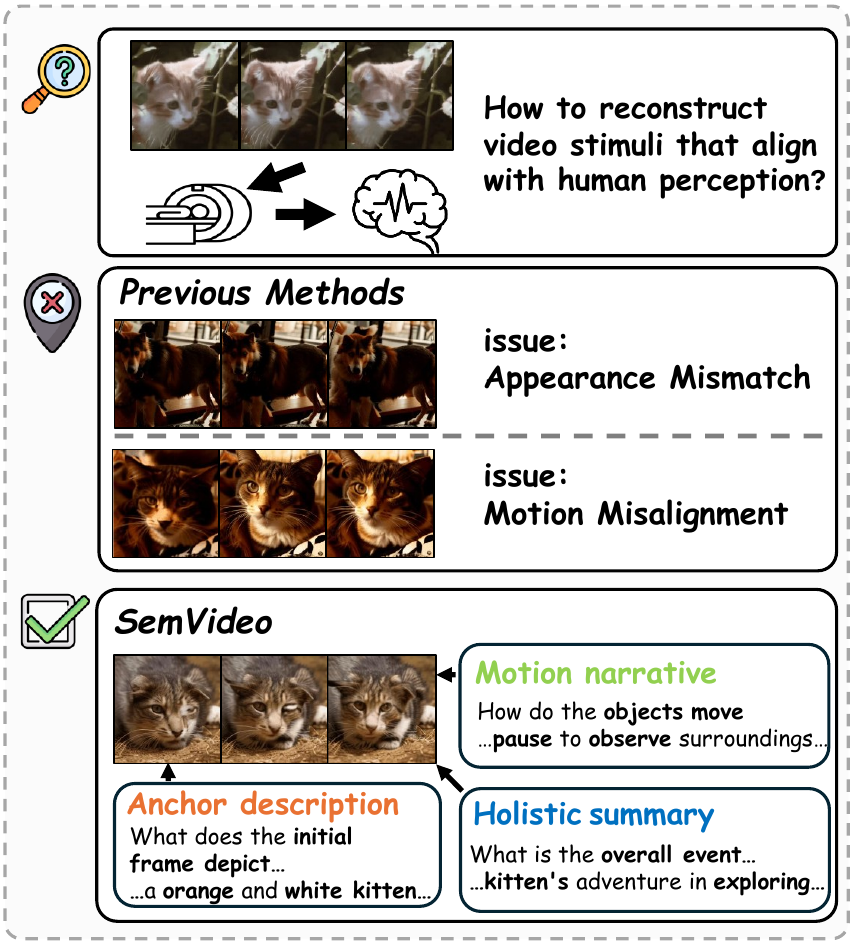}
    \vspace{-0.2cm}
    \caption{\textbf{Top:} While a subject watches an original video stimulus, their brain activity is recorded via fMRI. \textbf{Middle:} Reconstructed results from previous methods, which suffer from two issues: \textit{Appearance Mismatch and Motion Misalignment}. \textbf{Bottom:} Reconstructed result from SemVideo, achieves both semantic consistency (reconstructing the ``kitten'') and motion coherence (matching dynamic actions like ``crouching'' and ``turning'') by leveraging hierarchical semantic descriptions as intermediate targets to guide the fMRI signal decoding process.}
    \label{fig:overall}
\end{figure}
Understanding the information encoded in brain activity remains one of the central challenges in cognitive neuroscience. Reconstructing external stimuli from brain signals — particularly using non-invasive methods such as functional Magnetic Resonance Imaging (fMRI) — is an important yet highly challenging task. Neural decoding typically encompasses three fundamental tasks\cite{guo2025surveyfmriimagereconstruction}: classification, identification, and reconstruction. With the rapid advancement of deep learning techniques, prior work has made remarkable progress in both the classification\cite{Kamitani2005Decoding,Yargholi2016Brain} and identification\cite{Horikawa2015Generic,Kay2008Identifying} of static visual stimuli from fMRI data. Notably, the release of the Natural Scenes Dataset (NSD)\cite{Allen2022Massive}—the largest publicly available fMRI-image paired dataset — has catalyzed significant breakthroughs in image reconstruction\cite{Chen2024mindartist,Gong2025MindTuner,scottimindeye2}. Moreover, the emergence of powerful text-to-image diffusion models has brought high-quality, high-resolution image reconstruction from fMRI signals within reach\cite{wang2024mindbridge,chen2023seeing,Huo2024Neuropictor,Lu2023Minddiffuser,wang2024unibrain,Ozcelik2023Brain-diffuser,Quan2024Psychometry,Xia2024Dream,Xia2024Umbrae}, transforming what was once considered an unattainable goal into a practical possibility.


However, reconstructing \textit{dynamic visual stimuli }from fMRI signals remains a significant challenge due to the inherent limitations of fMRI, which relies on the slow hemodynamic response of BOLD signals. These signals integrate brain activity over several seconds, making it difficult to capture the rapid motion variations present in video stimuli. Consequently, fMRI-based video reconstruction becomes an even more complex and challenging task. Despite this, the development of video generation technology has greatly enhanced the feasibility of fMRI-based video reconstruction tasks\cite{chen2023cinematic,fosco2024ECCV,sun2024neurocine,gong2024neuroclips,lu2025animate}. However, two major issues remain unresolved in this task: (i) inconsistent visual representations of salient objects across frames, leading to appearance mismatches, and (ii) poor temporal coherence, resulting in motion misalignment or abrupt frame transitions.

In this work, inspired by neuroscience research indicating that the human brain perceives videos discretely\cite{Rabinoff2018Human,Scholler2012Toward} due to phenomena such as persistence of vision\cite{Ferry1892Persistence,Zhu2022Ultra-high,Chen2024Caring,Patel2015Persistence,Broegger2010Persistence} and delayed memory\cite{Logothetis2002Neural,Kim2012Biophysical,Ungerleider1998Neural}. This makes it impractical for brain signals to process every video frame and every pixel; instead, only keyframes elicit significant responses from the brain’s visual system. This suggests that focusing on key semantic perceptions, rather than exhaustive pixel-by-pixel analysis, aligns more closely with the efficiency and practicality of the human visual system.

To this end, we first propose \Semprompt, a module built upon a multimodal large language model (MLLM~\cite{qiao2025vthinkerinteractivethinkingimages,qiao2024wemathdoeslargemultimodal,qiao2025wemath20versatilemathbook,Tan2OCR}), which decomposes the original video stimuli into fine-grained, multi-level textual descriptions including holistic summaries, static anchor descriptions, and motion-oriented narratives. This hierarchical supervision better simulates the way humans process and recall visual experiences, addressing the semantic under-specification problem in traditional pipelines. 
Guided by this hierarchical supervision, we then present \Semvideo, a novel fMRI-to-video reconstruction framework guided by hierarchical semantic supervision and motion-aware latent decoding. Our framework is built upon three core components: 
(i) \textbf{Semantic Alignment Decoder.} To decode hierarchical semantic features from fMRI signals and adapt to the variable signal dimensions across different subjects, we propose the Semantic Alignment Decoder (SAD). This cross-subject multi-level semantic decoder consists of a subject-specific projector, a subject-shared mapper, and a Refineformer module, enabling precise semantic feature decoding from fMRI signals while minimizing noise.
(ii) \textbf{Motion Adaptation Decoder.} To reconstruct coherent action sequences from brain signals, we designed the Motion Adaptation Decoder (MAD). The core of the MAD is a tripartite attention fusion architecture that synergistically integrates spatial self-attention, temporal self-attention, and semantic-guided cross-attention mechanisms. By explicitly injecting semantic priors into the attention computation, this module effectively aligns motion latents with both spatial structures and semantic actions, enhancing the fidelity and continuity of the reconstructed video dynamics.
(iii) \textbf{Multi-Stage Semantic Fusion for Video Reconstruction}. We propose a multi-level, staged guidance strategy for conditional video generation. Instead of relying on a single semantic cue, we progressively condition the generation on three sources: static semantics (anchor descriptions), dynamic semantics (motion-oriented narratives), and holistic video-level semantics (holistic summaries). This structured conditioning ensures that the synthesized video aligns with both the static and temporal structure of the original perceptual input, yielding more coherent and realistic outputs.

Extensive experiments have demonstrated the superior performance and robustness of \Semvideo~on two widely-used public datasets, CC2017~\cite{Wen2017Neural} and HCP~\cite{marcus2011informatics}. We evaluate our method against previous approaches across three dimensions: semantic-level, pixel-level, and spatiotemporal-level metrics, achieving SOTA performance on 8 out of 10 metrics. Additionally, we provide a neuroscience-based interpretation of the effectiveness of \Semvideo~using ROI-wise visualization techniques, which confirm that our hierarchical semantic descriptions activate corresponding brain regions and validate the model's effectiveness.
\section{Related Work}
\label{sec:intro}
\subsection{Video Caption Generation}
To introduce richer semantic guidance,  a recent research~\cite{scottimindeye2,wang2024mindbridge,gu2022decoding,takagi2023high,scotti2024reconstructing,chen2023seeing,yang2026synmind} trend has shifted from unimodal visual feature alignment to dual-modal vision-text alignment. However, the effectiveness of this strategy is constrained by a significant bottleneck: the scarcity of high-quality, open-source video captioning models. Lacking dedicated video captioning tools, current research~\cite{fosco2024ECCV,gong2024neuroclips,lu2025animate} often resort to applying Image Captioning models, such as GIT~\cite{wang2022git}, BLIP~\cite{pmlr-v162-li22n,10.5555/3618408.3619222,xue2024xgenmmblip3familyopen} and SmallCap~\cite{ramos2022smallcap}), to discrete video frames. This method generates a series of short, static descriptions that fail to capture both the video's temporal dynamics and its fine-grained semantic details. Consequently, even when supervised by this textual modality, downstream reconstruction models are unable to acquire comprehensive information about either motion or semantics. This dual deficiency is a primary cause of persistent issues with motion coherence and semantic accuracy in generated videos, representing a core challenge for the field.

\subsection{Diffusion-based video generation}
The feasibility of leveraging the fine-grained semantic guidance for the reconstruction task is critically dependent on the recent advancements in video generation. 
Building on the remarkable progress in image synthesis with diffusion models~\cite{rombach2022high,ramesh2022hierarchical,ho2020denoising}, the field has rapidly advanced to text-to-video (T2V) diffusion models~\cite{Blattmann2023AlignYL,Singer2022MakeAV,Hong2022CogvideoL}.
These approaches, including 3D diffusion models~\cite{Ho2022video} and methods such as AnimateDiff~\cite{guo2024animatediff}, excel at synthesizing temporally coherent motion strongly conditioned by textual descriptions. 
This ability to translate semantic intent into dynamic visual content positions them as the ideal generative backbone for fMRI-to-video decoding tasks.
\subsection{fMRI-based Video Reconstruction}
Early approaches ~\cite{Wen2017Neural, Han2019Variational,Wang2022Reconstructing,Ganit2022Penny,Le2022Brain2Pix,Nishimoto2011Reconstructing} to video reconstruction treated the task as a sequence of independent image reconstructions. These methods employed Generative Adversarial Networks (GANs~\cite{Good2014GAN}) or Autoencoders (AEs~\cite{kingma2014auto}) to synthesize frames from brain signals, with the primary goal of recovering low-level visual features.
However, reconstructions from these models often lacked discernible semantic content.
In contrast, recent studies have largely shifted toward generative frameworks centered on diffusion models~\cite{guo2024animatediff,ho2022imagenvideo,Ho2022video}.
One line of work, including Brain Netflix~\cite{fosco2024ECCV} and MinD-Video~\cite{chen2023cinematic}, utilizes masked-brain modeling (MBM) to map fMRI signals into a unified latent space that drives a video diffusion model. 
Another class of methods~\cite{sun2024neurocine,gong2024neuroclips,lu2025animate} focus on aligning fMRI signals with the deep features from pretrained models-such as CLIP embeddings or VAE encodings of video frames-to guide video reconstruction. 
Despite significant improvements in quality, these advanced methods still exhibit two primary limitations: insufficient motion coherence and semantic deviation. This gap highlights a pressing need for a more fine-grained semantic supervision mechanism to guide the video generation process.
\section{Methodology}
\label{sec:method}
\begin{figure*}[htbp]
     \centering
     \includegraphics[width=1\textwidth, trim=0 0 0 0, clip]{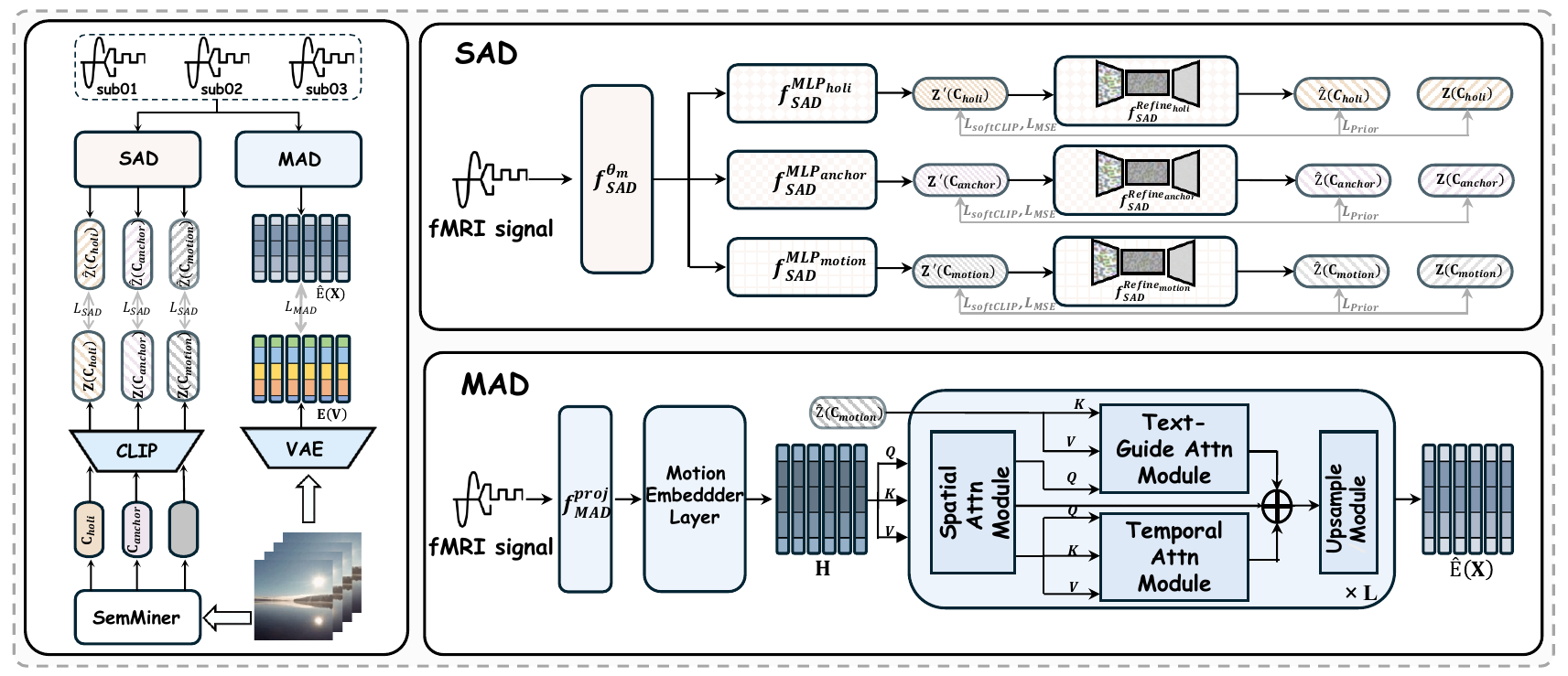}
     \vspace{-0.6cm}
     \caption{Overview of the \Semvideo~training pipeline. In the first stage, the Semantic Alignment Decoder is trained to map fMRI signals to three levels of semantic targets, denoted as $Z(C_L)$. In the second stage, the Motion Adaptation Decoder is trained to utilize the predicted motion semantics $\hat{Z}(C_{\text{motion}})$ to refine the latent embedding of each reconstructed frame.}
     \vspace{-0.1cm}
     \label{fig:framework}
 \end{figure*}
In this section, we present \Semprompt~and \Semvideo, two key components of our proposed fMRI-to-video reconstruction method. \Semprompt~deconstructs a given video into fine-grained, multi-level textual descriptions, organizing them into three hierarchical semantic perspectives: first-frame appearance, motion cues, and overall temporal consistency. \Semvideo~is a decoding framework that transforms fMRI signals into semantic embeddings and motion latents, which are then used as conditional inputs for a text-to-video generation model to reconstruct the original video stimuli.

\subsection{SemMiner}
To provide precise and diverse semantic supervision for fMRI-to-video reconstruction, we introduce \Semprompt, a hierarchical guidance module constructs three levels of semantic cues from the original video stimulus. Given a video stimulus $V$, these mined descriptions are structured into three complementary semantic perspectives. The \textbf{anchor description} $C_{\text{anchor}}$ captures the static visual content of the first frame, serving as a semantic anchor to ensure basic alignment between the reconstructed and original videos. The \textbf{motion-oriented narratives} $C_{\text{motion}}$ focus on fine-grained dynamic cues, particularly highlighting actions and dynamic transitions within the video. Finally, the \textbf{holistic summaries} $C_{\text{holi}}$ encapsulate a global semantic summary of the entire video, integrating both static and motion information into a coherent narrative. These descriptions establish a comprehensive framework, providing multi-level semantic guidance for the reconstruction.

\Semprompt~is implemented as a two-stage semantic decomposition module. In the first stage, it generates a simple semantic summary $C_{\text{basic}}$ of $V$ using a MLLM $\Psi$~\cite{zhang2025videollama3frontiermultimodal,zhang2024video,bai2025qwen25vl}, constrained to a maximum of 20 words, formulated as:
\begin{equation}
    C_{\text{basic}} = \Psi(P_{\text{basic}}, V),
\end{equation}
where $P_{\text{basic}}$ denotes the initial instruction used for synthesizing the basic description. This short summary serves as a ``rein'', to prevent the subsequent generation of diverse captions from becoming overly unconstrained or drifting semantically—analogous to controlling a ``runaway horse''~\cite{Long2024Multi-expert,Huang2025Survey,Zhou2023Least-to-Most,Khot2023Decomposed}. The imposed length limit ensures that $C_{\text{basic}}$ captures only the most essential content, thereby reducing the risk of introducing strong priors that could dominate later stages.

In the second stage, \Semprompt~is steered by carefully crafted instructions $P_{L}$ corresponding to specific semantic targets. Conditioned on the video $V$ and the initial summary $C_{\text{basic}}$, the framework elicits the target-specific descriptions:
\begin{equation}
    C_{L} {=} \Psi(P_{L}, C_{\text{basic}}, V), \quad L \in \{\text{anchor}, \text{motion}, \text{holi}\},
\end{equation}
where $P_{L}$ denotes the instruction specifically designed for each semantic objective. The detailed instructions used in \Semprompt~can be found in Appendix A.1.

\textbf{CC2017-SE}. To advance the fMRI-to-video reconstruction field, we introduce CC2017-SE, a semantic extension of the CC2017 dataset. Building upon the original dataset, we apply \Semprompt~to generate three t semantic descriptions — $C_{\text{anchor}}$, $C_{\text{motion}}$, and $C_{\text{holi}}$. We also provide quantitative and qualitative analysis of synthesized captions in Appendix A.2. The extended dataset is available at \url{https://huggingface.co/datasets/YanmHa/SemVideo}.

\subsection{SemVideo}
\begin{figure*}[htbp]
    \centering
    \includegraphics[width=1\textwidth, trim=0 0 0 0, clip]{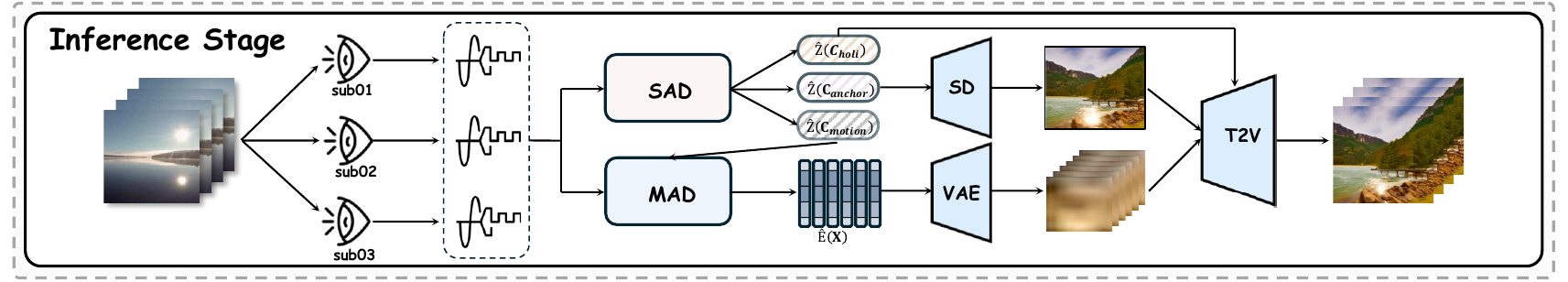}
    \vspace{-0.6cm}
     \caption{Overview of the \Semvideo~inference pipeline. fMRI signals are first decoded into $\hat{Z}(C_{\text{L}})$ by the SAD. $\hat{Z}(C_{\text{motion}})$ conditions the MAD to refine frame embeddings $\hat{E}(x)$, which are passed through a VAE decoder, generating a blurry video. $\hat{E}(x)$ and $\hat{Z}(C_{\text{anchor}})$ guide the SD model to generate anchor frame, combined with the blurry video and $\hat{Z}(C_{\text{holi}})$, is fed into a T2V model, yielding final reconstruction.}
     \vspace{-0.1cm}
     \label{fig:framework2}
\end{figure*}

\Semvideo~operates in three key stages, as illustrated in Figure.\ref{fig:framework} and Figure.\ref{fig:framework2}. First, the fMRI signals $X \in \mathbb{R}^{D_m}$ are decoded into semantic feature representations $\hat{Z}(C_{L}) \in \mathbb{R}^{77 \times 768}$, where $D_m$ denotes the number of activated voxels for subject $m$, and $Z(\cdot)$ represents the target feature space. Second, it reconstructs a motion-guided latent representation of the video, denoted as $\hat{E}(X)$, guided by the decoded motion semantics $\hat{Z}(C_{\text{motion}})$. Finally, the decoded semantic and motion features are fused to generate the reconstructed video via Conditional Video Render $\Phi$, formulated as:
\begin{equation}
\hat{V} = \Phi(\hat{Z}(C_{\text{anchor}}), \hat{Z}(C_{\text{holi}}), \hat E(X)).
\end{equation}
\noindent \textbf{Semantic Alignment Decoder (SAD)}
The objective of the Semantic Alignment Decoder (SAD) is to take the fMRI signal $X$ as input and output the predicted semantic feature $\hat{Z}(C_{L})$. 
Since the number of activated voxels varies across subjects, SAD first applies a subject-specific projection layer $f_{SAD}^{\theta_m}$ to project $X$ into a unified latent space of dimension $D_l$, yielding $X' = f_{SAD}^{\theta_m}(X) \in \mathbb{R}^{D_l}$. 
Next, a subject-shared encoder $f_{\text{SAD}}$ maps the normalized fMRI representation $X'$ to the CLIP text feature space, resulting in $\hat{Z}(C_{L}) = f_{\text{SAD}}(X')$. The encoder $f_{\text{SAD}}$ is composed of a four-layer multilayer perceptron ($f_{\text{SAD}}^{\textit{MLP}}$) followed by a Refineformer ($f_{\text{SAD}}^{\textit{Refine}}$), a causal transformer designed to maximize the extraction of meaningful neural activity while minimizing the influence of noise, thereby enhancing the alignment between fMRI signals and the target semantic space. The training of SAD is supervised by a combination of three training objectives:
\begin{equation}
    \mathcal{L}_\text{MSE} 
    = \bigl\|  Z(C_L) - f_{\text{SAD}}^{\textit{MLP}}(X') \bigr\|^2,
\end{equation}
\begin{equation}
\small
\begin{split} 
  \mathcal{L}_\text{SoftCLIP} 
  &= - \sum_{j=1}^B 
  \Biggl[
  \frac{\exp\Bigl(\frac{Z(C_L) \cdot Z(C_L^j)}{\tau}\Bigr)}
       {\sum_{k=1}^B \exp\Bigl(\frac{Z(C_L) \cdot Z(C_L^k)}{\tau}\Bigr)}
  \\ 
  &\qquad \times
  \log\Bigl(
       \frac{\exp\Bigl(\frac{f_{\text{SAD}}^{\textit{MLP}}(X') \cdot Z(C_L^j)}{\tau}\Bigr)} 
            {\sum_{k=1}^B \exp\Bigl(\frac{f_{\text{SAD}}^{\textit{MLP}}(X') \cdot Z(C_L^k)}{\tau}\Bigr)}
  \Bigr)
  \Biggr],
\end{split}
\end{equation}
\begin{equation}
\small
\mathcal{L}_{\text{refine}} \! = \! \mathbb{E}_{t \sim [1,T]}
    \left\|
      f_{\text{SAD}}^{\textit{Refine}}
      \! \left( Z_L^t, t, f_{\text{SAD}}^{\textit{MLP}}(X') \right)
      \! - \! Z(C_L)
    \right\|^2 \! , \label{eq:prior_loss}
\end{equation}
\begin{align}
\mathcal{L}_{\text{SAD}} 
&= \lambda_{\text{refine}}\, \mathcal{L}_{\text{refine}} 
   + \lambda_{\text{SoftCLIP}}\, \mathcal{L}_{\text{SoftCLIP}} 
   + \mathcal{L}_{\text{MSE}}, \label{eq:sfd_loss}
\end{align}
where $B$ denotes the batch size, $j$ indexes the $j$-th sample in the batch, and $\tau$ is a temperature hyperparameter. $T$ represents the total number of denoising timesteps, and $Z_L^t$ is the perturbed semantic embedding of $C_L$ at timestep $t$. $\lambda_{\text{refine}}$ and $\lambda_{\text{SoftCLIP}}$ are weighting coefficients that balance the contributions of the corresponding loss components.

\noindent \textbf{Motion Adaptation Decoder (MAD)} To improve the fidelity of motion dynamics reconstructed from brain signals, we propose a semantically guided motion adaptation decoding strategy. Let the original video stimulus be denoted as $V = \{v_1, v_2, \cdots, v_n\}$, where $n$ is the number of frames. We extract its motion latent representation using a pretrained VAE encoder $\mathbf{E}$, yielding $E(V) = \{e_1, e_2, \cdots, e_n\}$.

The objective of MAD is to take the fMRI signal $X$ as input and generate a sequence of predicted motion latents corresponding to the reconstructed video frames, denoted as $\hat{E}(X) = \{\hat e_1, \hat e_2, \cdots, \hat e_n\}$. To achieve this, MAD first projects the fMRI signal into the latent space using a subject-specific projection network $f_{\text{MAD}}^{\textit{proj}}$, implemented as a multilayer perceptron \footnote{The architecture of $f_{\text{MAD}}^{\textit{proj}}$ is detailed in Appendix C.1}. This projection yields a latent tensor $g = f_{\text{MAD}}^{\textit{proj}}(X) \in \mathbb{R}^{n \times D_{\text{emb}}}$, where $D_{\text{emb}}$ denotes the embedding dimension. Next, the tensor $g$ is processed by a Motion Embedder Layer to produce a sequence of  embeddings $S = \{s_1, s_2, \cdots, s_n\} \in \mathbb{R}^{n \times c \times h \times w}$, where $c \times h \times w$ is the dimension of the embedding latent space. This embedding sequence serves as the initial input to the subsequent attention-based decoding module. MAD conducts a hierarchical fusion attention mechanism: 

\noindent(i) To capture intra-frame structure, we apply \textit{spatial attention} to each $s_i$. The latent tensor is reshaped and attended via scaled dot-product attention:
\begin{equation}
    \hat{S} = \text{Reshape}(S) \in \mathbb{R}^{(n \cdot h \cdot w) \times c},
\end{equation}
\begin{equation}
    E_{\text{spat}} = \text{Softmax}\left(\frac{Q_{\text{spat}} K_{\text{spat}}^\top}{\sqrt{d}}\right) V_{\text{spat}},
\end{equation}
where $
    Q_{\text{spat}} = \hat{S} W_{Q_{\text{spat}}},K_{\text{spat}} = \hat{S} W_{K_{\text{spat}}},V_{\text{spat}} = \hat{S} W_{V_{\text{spat}}}.$ (ii) To model inter-frame dependencies, we apply \textit{temporal attention} across the temporal axis. We first flatten each spatially attended frame into a temporal sequence:
\begin{equation}
    \tilde{S} = \text{Reshape}(E_{\text{spat}}) \in \mathbb{R}^{(c \cdot h \cdot w) \times n},
\end{equation}
The attention computation follows the same formulation as in the spatial attention and is omitted here for brevity. 

\noindent(iii) We further incorporate \textit{semantic guided cross-attention}. Let $\hat Z(C_{\text{motion}})$ be the predicted motion descriptions from the video, as formulated by:
\begin{equation}
    E_{\text{cross}} = \text{Softmax}\left(\frac{Q_{\text{cross}} K_{\text{cross}}^\top}{\sqrt{d}}\right) V_{\text{cross}},
\end{equation}
where $Q_{\text{cross}} = \tilde{S}  W_{Q_{\text{cross}}}, K_{\text{cross}} = \hat Z(C_{\text{motion}}) W_{K_{\text{cross}}}$ and $V_{\text{cross}} = \hat Z(C_{\text{motion}}) W_{V_{\text{cross}}}$. Finally, the output of MAD can be represented as:
\begin{equation}
{\hat e}_{i} = \lambda_{\text{spat}}\,{e}_i^{\text{spat}}
           \;+\;\lambda_{\text{temp}}\,{e}_i^{\text{temp}}
           \;+\;{e}_i^{\text{cross}},
\end{equation}
where $\lambda_{\text{spat}}$ and $\lambda_{\text{temp}}$ are weights for ${e}_{\text{spat}}$ and ${e}_{\text{temp}}$, respectively. The training objectives of MAD are formulated by:
\begin{align}
\mathcal{L}_{MAD} = & \frac{1}{n} \sum_{i=1}^{n} \left| \hat e_i - e_i \right| \nonumber \\
& - \frac{1}{2 n} \sum_{j=1}^{n} \log \frac{\exp \left( \text{sim}(\hat e_j, e_j) / \tau \right)}{\sum_{k=1}^{n} \exp \left( \text{sim}(\hat e_j, e_k) / \tau \right)} \nonumber \\
& - \frac{1}{2 n} \sum_{j=1}^{n} \log \frac{\exp \left( \text{sim}(e_j, \hat e_j) / \tau \right)}{\sum_{k=1}^{n} \exp \left( \text{sim}(e_j, \hat e_k) / \tau \right)}
\end{align}
where $\text{sim}(\cdot, \cdot)$ is cosine similarity. 

\noindent \textbf{Conditional Video Render (CVR)} To fully leverage the information decoded from fMRI signals, we design CVR — a sequential inference framework that reconstructs the video by progressively integrating the fMRI-derived cues at each stage of the generation process.
First, the decoded motion latents  
$\hat{E}(X)=\{\hat{e}_1,\hat{e}_2,\dots,\hat{e}_n\}$  
are passed through the pretrained VAE decoder $\mathbf{D}$ to produce a sequence of motion frames  
$\{I_i^{\text{motion}}\}_{i=1}^{n}$.  
Next, the decoded anchor feature $\hat{Z}(C_{\text{anchor}})$ is combined with the first motion frame $I_1^{\text{motion}}$ and fed into a text-to-image (T2I) model, yielding the initial reconstructed frame $\hat{v}_1$.  
Finally, a pretrained text-to-video (T2V) generator is steered jointly by three sources:  
(i) the holistic semantic guidance $\hat{Z}(C_{\text{holi}})$,  
(ii) the anchor frame $\hat{v}_1$, and  
(iii) the motion-frame sequence $\{I_i^{\text{motion}}\}_{i=1}^{n}$.  
This combination encourages both temporal smoothness and semantic fidelity, enabling CVR to synthesize a coherent video that closely matches the original visual stimulus inferred from the fMRI signals.
\definecolor{sigextreme}{RGB}{220, 234, 247}

\definecolor{sigvery}{RGB}{254, 241, 243}

\definecolor{sigcolor}{RGB}{254, 246, 190}

\definecolor{notsigcolor}{RGB}{232, 243, 225}

\newcommand{\sigvery}[1]{\cellcolor{sigvery}#1}
\newcommand{\sigextreme}[1]{\cellcolor{sigextreme}#1}
\newcommand{\sig}[1]{\cellcolor{sigcolor}#1}
\newcommand{\notsig}[1]{\cellcolor{notsigcolor}#1}
\section{Experiments}
\begin{table*}[htbp]
    \setlength{\belowcaptionskip}{3pt}
	\centering
    \renewcommand{\arraystretch}{1.2}
	\resizebox{\textwidth}{!}{
		\begin{tabular}{l|ccccc|ccc|cc}
			\toprule
			\multirow{2}{*}{Method} & \multicolumn{5}{c|}{Semantic-Level} & \multicolumn{3}{c|}{Pixel-Level} & \multicolumn{2}{c}{ST-Level} \\
			& 2-way-I$\uparrow$ & 2-way-V$\uparrow$ & 50-way-I$\uparrow$& 50-way-V$\uparrow$ & VIFI-score$\uparrow$ & SSIM$\uparrow$ & PSNR$\uparrow$ & Hue-pcc$\uparrow$ & CLIP$\uparrow$ & EPE $\downarrow$ \\
			\midrule
			Nishimoto~\cite{Nishimoto2011Reconstructing}(Current Biology'11) & \sigextreme{0.742} & -- & -- & -- & -- & \sigextreme{0.119} & \sigextreme{8.383} & \sigextreme{0.737} & -- & -- \\
			Wen~\cite{Wen2017Neural}(Cerebral Cortex'17) & \sigextreme{0.771} & \sigextreme{0.070} & -- & \sigextreme{0.166} & -- & \sigextreme{0.130} & \sigextreme{8.301} & \sigextreme{0.637} & -- & -- \\
			Kupershmidt~\cite{Ganit2022Penny}(arXiv'22) & \sigextreme{0.769} & \sigextreme{0.768} & \sigextreme{0.179} & -- & \sigextreme{0.591}& \sigextreme{0.108} & \sigextreme{\underline{11.030}} & \sigextreme{0.583} & \sigextreme{0.399} & \sigextreme{6.344} \\
            f-CVGAN~\cite{Wang2022Reconstructing}(Cerebral Cortex'22) & \sigextreme{0.777} & \sigextreme{0.721} & -- & -- & \sigextreme{0.592} & \sigextreme{0.108} & \sigextreme{\textbf{11.430}} & \sigextreme{0.583} & \sigextreme{0.399} & \sigextreme{6.344} \\
			Mind-video~\cite{chen2023cinematic}(NeurIPS'23) & \sigextreme{0.797} & \notsig{0.848} & \sigextreme{0.174} & \sigextreme{0.197} & \sigextreme{0.593}& \sigextreme{0.177} & \sigextreme{8.868} & \sigextreme{0.768} & \sigextreme{0.409} & \sigextreme{6.125} \\
			NeuroClips~\cite{gong2024neuroclips}(NeurIPS'25) & \sigextreme{0.806} & \sig{0.834} & \sigextreme{0.203} & \sigextreme{0.220} & \notsig{\underline{0.602}}& \sigextreme{\textbf{0.390}} & \sigvery{9.211} & \sigextreme{0.812} & \sigextreme{\underline{0.513}} & \sigextreme{4.833} \\
            Mind-Animator~\cite{lu2025animate}(ICLR'25)& \sigextreme{0.805} & \sig{0.830} & \sigextreme{0.192} & \sigextreme{0.186}& \notsig{\textbf{0.608}} & \notsig{\underline{0.321}} & \sigextreme{9.220} & \sigextreme{0.786} & \sigextreme{0.425} & \sigextreme{5.422} \\
            NEURONS~\cite{wang2025neurons}(ICCV'25)& \sigextreme{\underline{0.815}} & \sig{\underline{0.853}} & \sigextreme{\underline{0.213}} & \sigextreme{\underline{0.246}}& \sigextreme{0.597} & \sigextreme{0.285} & \notsig{9.526} & \sigextreme{\underline{0.830}} & \sigextreme{0.482} & \sigextreme{\underline{4.827}} \\
            \cdashline{1-11}           
			Ours & \textbf{0.838} & \textbf{0.865} & \textbf{0.225} & \textbf{0.264} & \textbf{0.608} & \underline{0.321} & 9.448 & \textbf{0.849} & \textbf{0.526} & \textbf{4.788} \\
			\bottomrule
		\end{tabular}
	}
    \vspace{-0.1cm}
    \caption{Quantitative comparison of reconstruction results on the CC2017 dataset. All metrics are averaged across all samples and three subjects, with the best results highlighted in bold and the second-best results underlined. 100 sample sets were constructed using the bootstrap method for hypothesis testing. Colors reflect statistical significance compared to our model. $p < 0.0001$(\textcolor[RGB]{155, 187, 225}{blue}); $p < 0.01$(\textcolor[RGB]{251, 171, 171}{pink}); $p < 0.05$(\textcolor[RGB]{239, 227, 141}{yellow}); $p > 0.05$(\textcolor[RGB]{156, 203, 127}{green})}.
    \label{tab:main_quantitative}
    \vspace{-0.1cm}
\end{table*}

\begin{figure*}[htbp]
    \centering
    \includegraphics[width=0.98\textwidth, trim=0 0 0 0, clip]{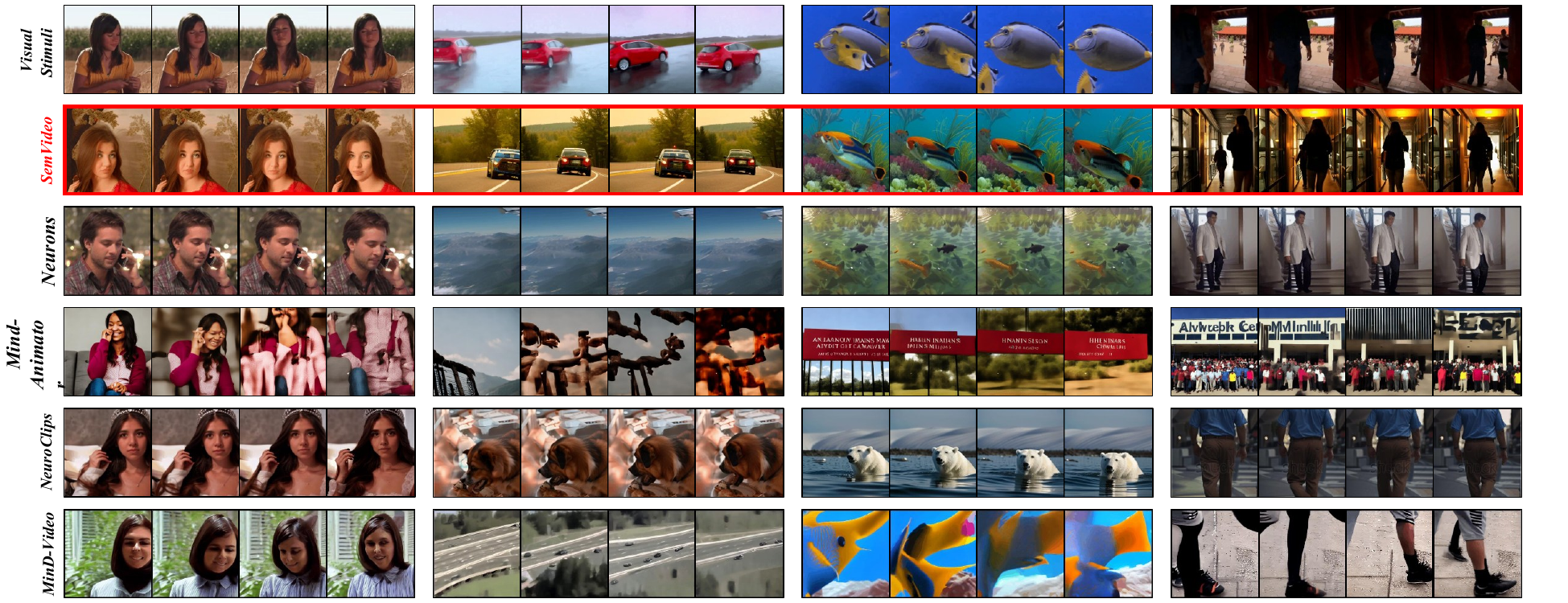}
    \vspace{-0.1cm}
    \caption{Qualitative comparison of reconstruction results on the CC2017 dataset between previous methods and our proposed \Semvideo. Reconstructions generated by \Semvideo are highlighted with red boxes.}
    \label{fig:qualitative}
\end{figure*}

\begin{table}[htbp]
    \centering
    \renewcommand{\arraystretch}{1.1}
    \vspace{-0.3cm}
    \resizebox{0.48\textwidth}{!}{ 
        \begin{tabular}{l|cc|cc|cc}
            \toprule
            \multirow{2}{*}{Method} & \multicolumn{2}{c|}{Semantic} & \multicolumn{2}{c|}{Pixel} & \multicolumn{2}{c}{ST-Level} \\
             & 2-way-I$\uparrow$ & 2-way-V$\uparrow$ & PSNR$\uparrow$ & Hue-pcc$\uparrow$ & CLIP $\uparrow$ & EPE $\downarrow$ \\
            \midrule
            Nishimoto & \sigextreme{0.658} & -- & \sigextreme{\textbf{11.316}} & \sigextreme{0.645} & -- & -- \\
            Wen & \sigextreme{0.702} & -- & \sigextreme{10.197} & \sigextreme{0.727 }& -- & -- \\
            Mind-video & \sigextreme{0.779} & \sigvery{0.769} & \notsig{9.275} & \sigextreme{0.793} & \sigextreme{0.499} & \sigextreme{9.29 }\\
            Mind-Animator & \sigextreme{0.786}   & \sig{0.778} & \sigextreme{11.233} & \sigextreme{0.829} & \sig{0.511} & \sigextreme{7.08} \\
            \cdashline{1-7}
            \textbf{Ours} & \textbf{0.815} & \textbf{0.796} & 9.336 & \textbf{0.858} & \textbf{0.523} & \textbf{4.05} \\
            \bottomrule
        \end{tabular}
    }
    \caption{Quantitative comparison on the HCP dataset.}
    \label{tab:main_quantitative_hcp}
\end{table}
\textbf{Datasets and Pre-processing}: We evaluate our method on two publicly available fMRI-video datasets: CC2017~\cite{Wen2017Neural} and a subset of the HCP 7T dataset~\cite{marcus2011informatics}. 
\textbf{CC2017} contains fMRI recordings from three subjects (3T MRI) while watching 23 high-definition natural movie clips (8 minutes each, 30 FPS). To improve signal-to-noise ratio (SNR), training clips are repeated twice and test clips ten times; the fMRI responses are averaged accordingly. After aligning video frames with the fMRI temporal resolution (TR = 2s), we obtain 8,640 training samples and 1,200 test samples. Visually responsive voxels are selected by computing the temporal correlation between repeated presentations, applying Fisher's z-transform, and conducting a one-sample t-test (Bonferroni corrected, $p < 0.05$). This yields 13,447; 14,828; and 9,114 voxels for the three subjects, respectively. A 4-second BOLD lag is applied to compensate for the hemodynamic response delay. \textbf{HCP 7T}: We use a subset of the HCP 7T dataset, selecting three subjects (IDs: 100610, 102816, 104416) following prior work. The data undergoes HCP’s preprocessing pipeline, including motion/distortion correction, high-pass filtering, ICA denoising, and MNI registration. We extract 5,820 voxels from~\cite{glasser2016multi}. We partition clips into training and test sets with a 9:1 split. A 4-second hemodynamic lag is similarly applied.

\begin{table*}[htbp]
	\centering
    \renewcommand{\arraystretch}{1.2} 
	\resizebox{\textwidth}{!}{
		\begin{tabular}{l|ccccc|ccc|cc}
			\toprule
			\multirow{2}{*}{Method} & \multicolumn{5}{c|}{Semantic-Level} & \multicolumn{3}{c|}{Pixel-Level} & \multicolumn{2}{c}{ST-Level} \\
			& 2-way-I$\uparrow$ & 2-way-V$\uparrow$ & 50-way-I$\uparrow$ & 50-way-V$\uparrow$ & VIFI-score$\uparrow$ & SSIM$\uparrow$ & PSNR$\uparrow$ & Hue-pcc$\uparrow$ & CLIP $\uparrow$ & EPE $\downarrow$ \\
			\midrule
			Ours (full) & \textbf{0.826} & \textbf{0.860} & \textbf{0.211} & \textbf{0.239} & \textbf{0.590} & 0.330 & \textbf{9.626} & \textbf{0.841} & \textbf{0.502} & \textbf{4.768} \\
			\ w/o $C_{anchor}$ & \sigextreme{0.786} & \sigextreme{0.808} & \sigextreme{0.158} & \sigextreme{0.147} & \sigextreme{0.534} & \sigextreme{ \textbf{0.361}} & \sigvery{9.454} & \sigvery{0.835} & \notsig{0.488} & \notsig{4.796} \\
			\ w/o $C_{holi}$ & \sig{0.819} & \sig{0.849} & \sigvery{0.205} & \notsig{0.221} & \notsig{0.584} & \sigextreme{0.299} & \sigvery{9.492} & \sigvery{0.834} & \sigvery{0.490} & \sigvery{4.859} \\
			\ w/o $C_{motion}$ & \notsig{0.818} & \sigextreme{0.846} & \sigextreme{0.207} & \sigextreme{0.216} & \sigvery{0.583} & \sigextreme{0.285} & \sigextreme{9.353} & \sigextreme{0.741} & \sigextreme{0.481} & \sigextreme{4.930} \\
			\bottomrule
		\end{tabular}
	}
    \vspace{-0.1cm}
    \caption{Ablation study evaluating the impact of removing different semantic descriptions—$C_{anchor}$, $C_{holi}$, and $C_{motion}$—from the decoding targets of the Semantic Alignment Decoder (SAD). All results are reported from subj01 of the CC2017 dataset.}
    \vspace{-0.3cm}
    \label{tab:ablation}
\end{table*}

\noindent\textbf{Evaluation Metrics }: We follow previous works that introduce the evaluation metrics at three complementary levels: \noindent\textbf{Semantic-Level}:  
(i) \emph{N-way top-1 retrieval} in four settings—2-way/50-way, frame-level~\cite{dosovitskiy2021image}]/video-level~\cite{Tong2022VideoMAE}.  
(ii) \emph{VIFI}~\cite{rasheed2023fine}: cosine similarity of VIFICLIP features between ground-truth and reconstructions.
\noindent\textbf{Pixel-Level}:  SSIM, PSNR, and Hue-PCC~\cite{Swain1991color} quantify low-level fidelity.
\noindent\textbf{Spatiotemporal-Level (ST-Level)}:  
(i) \emph{CLIP-PCC}: average cosine similarity of CLIP embeddings of adjacent frames; set to 0 when VIFI\,$<0.6$ to avoid spurious inflation.  
(ii) \emph{EPE}~\cite{Barron1994Performance}: quantifies motion fidelity by calculating the average Euclidean distance between predicted and ground-truth optical flow vectors at each pixel.

\noindent\textbf{Implementation}:
The training strategy of the experiment involves pre-training across all subjects, followed by fine-tuning on a single subject. Consistent with prior research~\cite{gong2024neuroclips}, we adopt AnimateDiff for T2V task. All models are trained for 100 epochs with AdamW. Loss weights are $\lambda_{\text{SoftCLIP}}{=}0.1$ and $\lambda_{\text{refine}}{=}0.5$. Please refer to Appendix C.1 for more implementation details.

\subsection{Main Results}
We report the quantitative performance of our method on the CC2017 dataset in Table.\ref{tab:main_quantitative} and qualitative comparisons in Figure.\ref{fig:qualitative}. \Semvideo~achieves remarkable and comprehensive performance across all evaluation levels.

\noindent \textbf{Superior Semantic Accuracy and Fidelity.} Our model outperforms NeuroClips~\cite{gong2024neuroclips} and Neurons~\cite{wang2025neurons}, achieving a 2-way-V score of 0.865 and a 50-way-V score of 0.264, demonstrating superior semantic consistency and discriminability. On the VIFI-score, which reflects overall semantic-video alignment, \Semvideo~ties with Mind-Animator\cite{lu2025animate} at 0.608, indicating strong semantic fidelity. This quantitative superiority is also reflected visually in Figure.\ref{fig:qualitative} where the reconstructions of \Semvideo~are more aligned with the ground truth video stimuli, accurately capturing central objects like “fish” and “cars”.

\noindent \textbf{Competitive Low-Level Reconstruction Quality.}
For pixel-level reconstruction quality, our model achieves the highest hue-pcc (0.849) and reaches a competitive SSIM of 0.321 and PSNR of 9.448, closely trailing the SoTA method. This high fidelity is also evident visually, as it effectively recovers key low-level features such as lighting and shadow contrasts (e.g., rightmost column in Figure.\ref{fig:qualitative}).

\noindent \textbf{Enhanced Spatial-Temporal Coherence.}
Our model achieves the highest CLIP similarity (0.526), and the lowest EPE (4.788), highlighting better preservation of spatial-temporal coherence and alignment with high-level perceptual features. Most notably, this is demonstrated by \Semvideo's ability to excel at modeling temporal dynamics, successfully reconstructing coherent motion patterns—such as ``a person turning their head'' (Figure.\ref{fig:qualitative},  first column)—which remain challenging for previous methods. These results underscore \Semvideo’s strength in preserving both spatial detail and temporal consistency in fMRI-to-video reconstruction.

\noindent \textbf{Strong Generalization Ability.}
To assess generalization, we further evaluate on the HCP dataset. As shown in Table.\ref{tab:main_quantitative_hcp}, \Semvideo~maintains strong performance, demonstrating its robustness across subjects and datasets.

\begin{figure}[htbp]
    \includegraphics[width=0.98\linewidth, trim=0 0 0 0, clip]{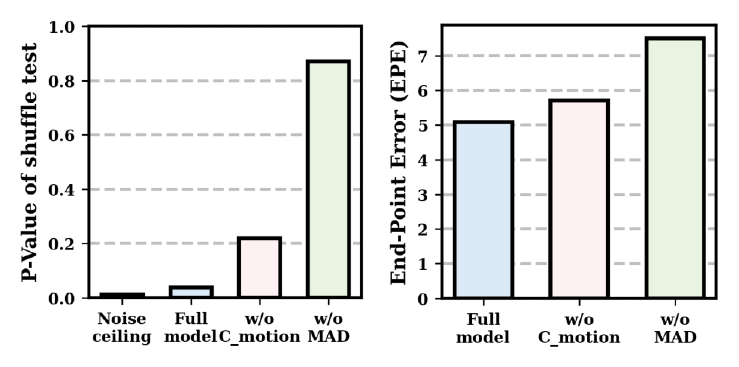}
    \vspace{-0.5cm}
    \caption{The figure presents two sets of results from subj01 of the CC2017 dataset. On the left, the results of a shuffle test are shown. On the right, the results of an ablation study are displayed, evaluating the impact of the $C_{motion}$ and $MAD$ guidance components.}
    \label{fig:ablation}
\end{figure}

\begin{figure*}[htbp]
    \centering
    \includegraphics[width=0.98\textwidth, trim=0 0 0 0, clip]{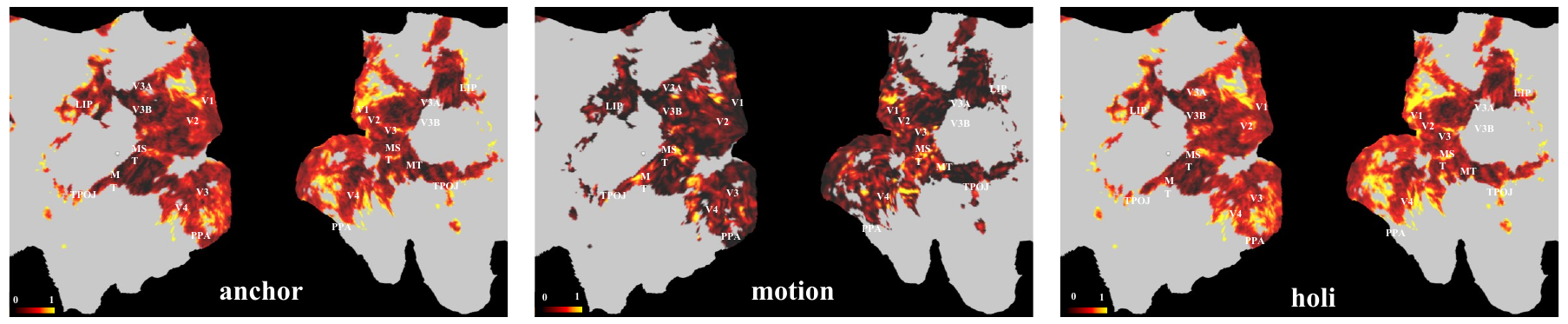}
    \vspace{-0.3cm}
    \caption{Importance visualization of different ROIs of subject 1 on CC2017, based on the fitted weights from the first layer of the SAD. Weights from each module are averaged and normalized to the $[0, 1]$ range for comparison.  Please refer to Appendix D for
more subjects’ importance visualization.}
    \label{fig:cortex}
\end{figure*}
\subsection{\textbf{Validating the Source of Motion Improvement.}}
Following the work of~\cite{lu2025animate}, we reconstruct video clips using Tune-a-video~\cite{wu2023tune} which lacks motion priors and conduct shuffle tests to further validate the contributions of both the $MAD$ and the motion-oriented narratives $C_{motion}$ to motion improvement. 
Specifically, for each 12-frame reconstructed video, we randomly shuffle the order of the frames 100 times and compute the EPE on both the original and shuffled frames. 
We estimate the P-value using the following formula: $P = \sum_{i=1}^{100} \delta_i / 100$, where $\delta_i = 1$ if the $i^{th}$ shuffle outperforms the reconstruction result in the original order based on the metrics, and $\delta_i = 0$ otherwise.
A lower P-value signifies a closer alignment between the sequential order of the reconstructed video and the ground truth. 
It can be observed that the P-value for our full model is significantly low, confirming that its predicted inter-frame order robustly matches the ground truth, but rises substantially when $C_{motion}$ is removed and spikes to 0.84 when the entire $MAD$ is ablated, as illustrated in Figure.\ref{fig:ablation} left.
This indicates that we have indeed decoded motion information from $MAD$ and fMRI-derived $C_{motion}$. 
This conclusion is reinforced by a comparison of the EPE score, as shown in Figure.\ref{fig:ablation} right. The EPE increase when removing $C_{motion}$ and the spike when ablating the $MAD$ confirm the crucial role of motion semantic guidance and that the $MAD$, not T2V priors, is the source of decoded motion improvement.
\subsection{Ablation Study}
In this section, we conduct a detailed ablation study to evaluate the individual contributions of the $C_{anchor}$, $C_{holi}$, and $C_{motion}$ components to our \Semvideo. The quantitative results, presented in Table.\ref{tab:ablation}, unequivocally demonstrate that the removal of any single component leads to a performance degradation across nearly all metrics, affirming their indispensability.
Specifically, the exclusion of $C_{motion}$ causes the most precipitous decline in pixel-level and ST-level metrics. The absence of $C_{anchor}$, in contrast, significantly impairs semantic-level fidelity. The $C_{holi}$ proves integral for overall performance, as its removal adversely affects all metrics, underscoring its global guidance under video reconstruction.
Crucially, and in contrast to prior works employing conventional latent decoders, the removal of $C_{motion}$ also results in a notable drop in semantic metrics. This key finding substantiates the efficacy of our multi-level semantic guidance mechanism. It demonstrates that our motion latent space is not only responsible for ensuring pixel-level consistency and temporal smoothness but is also endowed with semantic understanding, thereby guaranteeing the coherence of the synthesized actions.


\subsection{Neuroscience Interpretability}
To investigate the neural interpretability of our model, we generated voxel-wise importance maps over the cerebral cortex. The results, presented in Figure.\ref{fig:cortex}, shed light on the neural substrates underlying each core component.
The anchor component (Figure.\ref{fig:cortex}, left) predominantly draws from a hierarchy of visual regions, with a notable emphasis on higher-level visual cortices, confirming its role in encoding abstract scene structure. The motion component (Figure.\ref{fig:cortex}, middle) aligns well with established motion-processing regions, including MT, MST, and TPOJ~\cite{Born2005}. Interestingly, it also shows strong activations in V1, supporting the existence of parallel, non-hierarchical pathways—where V1 may directly convey motion signals to MT~\cite{Nassi2009}—in contrast to the traditional view of a strictly feedforward dorsal stream~\cite{Zeki1988}. Lastly, the holistic component (Figure.\ref{fig:cortex}, right) exhibits a distributed pattern of importance across both visual and motion-related areas. This balanced integration highlights its function as a global information aggregator, responsible for synthesizing perceptual and temporal features to support coherent reconstruction. This hierarchical semantic mapping enables \Semvideo~to generate more perceptually faithful reconstructions and similar activation distributions across subjects demonstrate stable feature capture for both static semantics and dynamic behaviors.
\section{Conclusion}
Reconstructing dynamic visual stimuli from brain activity remains a challenge in cognitive neuroscience. In this work, we propose \Semvideo, an innovative fMRI-to-video reconstruction framework featuring a novel semantic alignment decoder, a motion adaptation decoder and a conditional video render. By leveraging multi-level semantic cues, \Semvideo~effectively addresses the challenges of both semantic sparsity and temporal coherence in video reconstruction. Our experimental evaluation demonstrates the superior performance of \Semvideo~on two public datasets, surpassing existing methods across multiple metrics, including semantic, pixel, and spatiotemporal levels. Furthermore, our use of ROI-wise visualization techniques provides valuable neuroscience-driven insights into the effectiveness of the proposed method. We believe \Semvideo~will set a new SOTA in fMRI-to-video decoding and lay the foundation for future brain-based video reconstruction.

\section*{Acknowledgement}
This work was supported in part by the Beijing Natural Science Foundation under Grant QY25342; in part by
the National Natural Science Foundation of China under Grant 62502045; in part by the Postdoctoral Fellowship Program of CPSF under Grant GZC20251088.

{
    \small
    \bibliographystyle{ieeenat_fullname}
    \bibliography{main}
}
\clearpage

%
\definecolor{cvprblue}{rgb}{0.21,0.49,0.74}

\def\confName{CVPR}
\def\confYear{2026}
\appendix

\section{SemMiner}
\subsection{The Implementation Details}
\label{Sec:appendix_semprompt_implementation}
Since the objective of \Semprompt~is to utilize a Multimodal Large Language Model (MLLM) to generate multi-level semantic descriptions of a video from diverse perspectives, directly prompting the MLLM for this complex task often results in hallucinations or inaccurate content. To address this, we decompose the overall objective into a sequence of ordered subtasks—forming a prompt chain—that guides the MLLM progressively from simpler prompts to more complex ones. This step-by-step prompting strategy enhances the model’s reasoning ability and output accuracy, while significantly reducing the risk of generating false or unreliable information~\cite{Long2024Multi-expert,Huang2025Survey,Zhou2023Least-to-Most,Khot2023Decomposed}.

\begin{figure*}[htbp]
    \centering
    \includegraphics[width=0.98\textwidth, trim=0 10 0 10, clip]{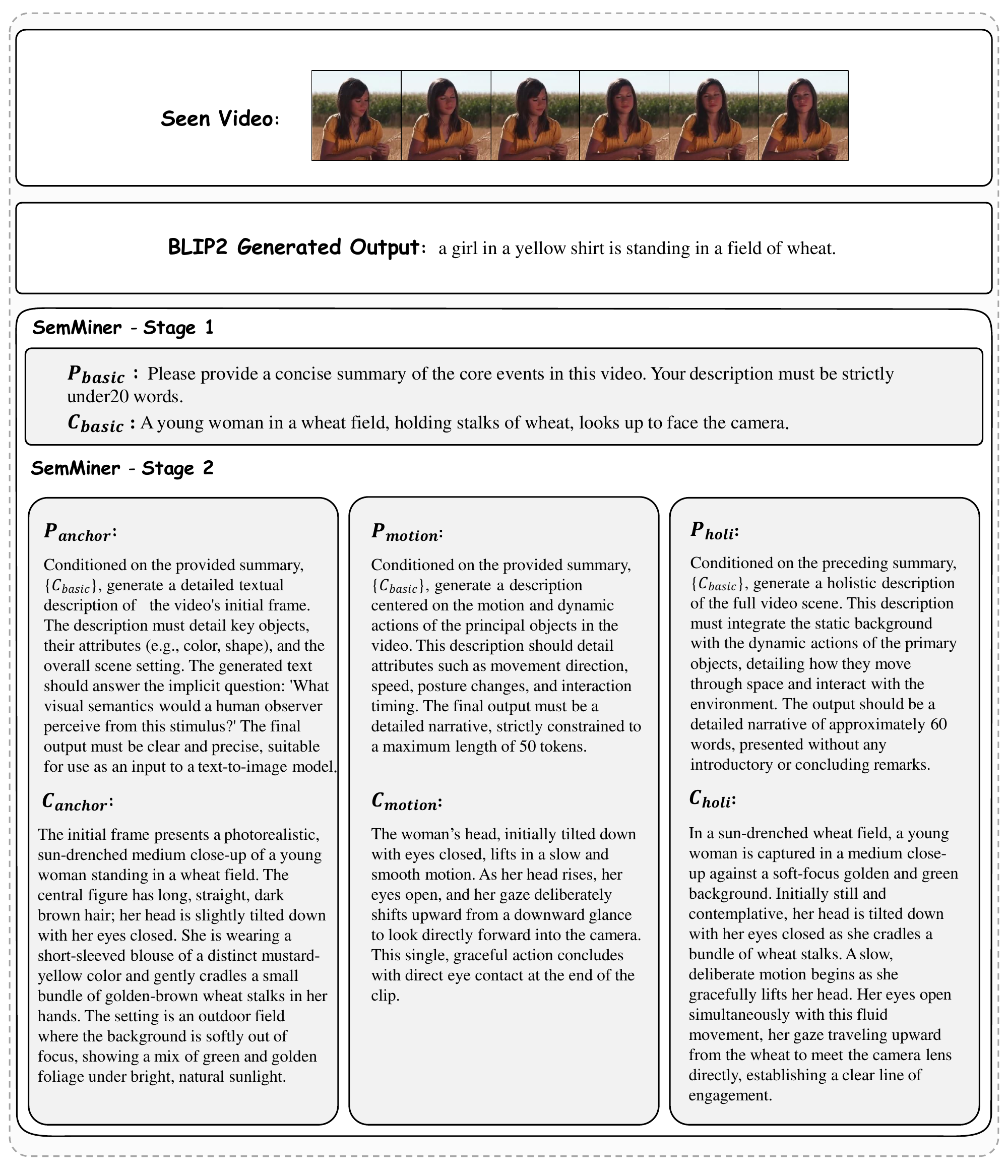}
    \vspace{-0.3cm}
    \caption{Prompt templates used in \Semprompt~ and qualitative comparison of captions generated by BLIP2 and \Semprompt. \Semprompt~ employs four distinct prompt templates to generate hierarchical semantic descriptions for a given video stimulus. The process begins with the Basic Prompt $P_{basic}$, which establishes a core semantic summary to anchor subsequent generations. This is followed by three specialized prompts in stage 2: (i) $P_{anchor}$, which produces a detailed description of the initial frame, optimized for guiding text-to-image models; (ii) $P_{motion}$, which focuses on extracting dynamic information such as object motion, direction, speed, and posture changes; and (iii) $P_{holi}$, which integrates static and dynamic elements to form a coherent, high-level narrative of the entire video. Compared to BLIP2, which tends to generate short and static descriptions, \Semprompt~ provides richer and more perspective-specific captions.}
    \label{fig:prompt}
\end{figure*}

\Semprompt~adopts a two-stage pipeline prompting strategy. In Stage 1, a basic prompt $P_{basic}$ is used to produce a concise core event summary $C_{basic}$, which serves as an anchoring ``rein'' to constrain and guide subsequent generation. Without this focused summary, the following caption generation process is prone to semantic drift. In Stage 2, \Semprompt~uses three specialized prompts in conjunction with $C_{basic}$ to generate three decoupled semantic descriptions: a static anchor description ($C_{anchor}$) of the initial frame, a motion-focused narrative ($C_{motion}$) describing the dynamic content, and a holistic summary ($C_{holi}$) that integrates both static and dynamic elements into a coherent spatio-temporal account. The specific prompt templates used in our experiments are illustrated in Figure.\ref{fig:prompt}.

\subsection{The Results of SemPrompt}
\begin{figure*}[htbp]
    \centering
    \includegraphics[width=0.98\textwidth, trim=130 80 90 80, clip]{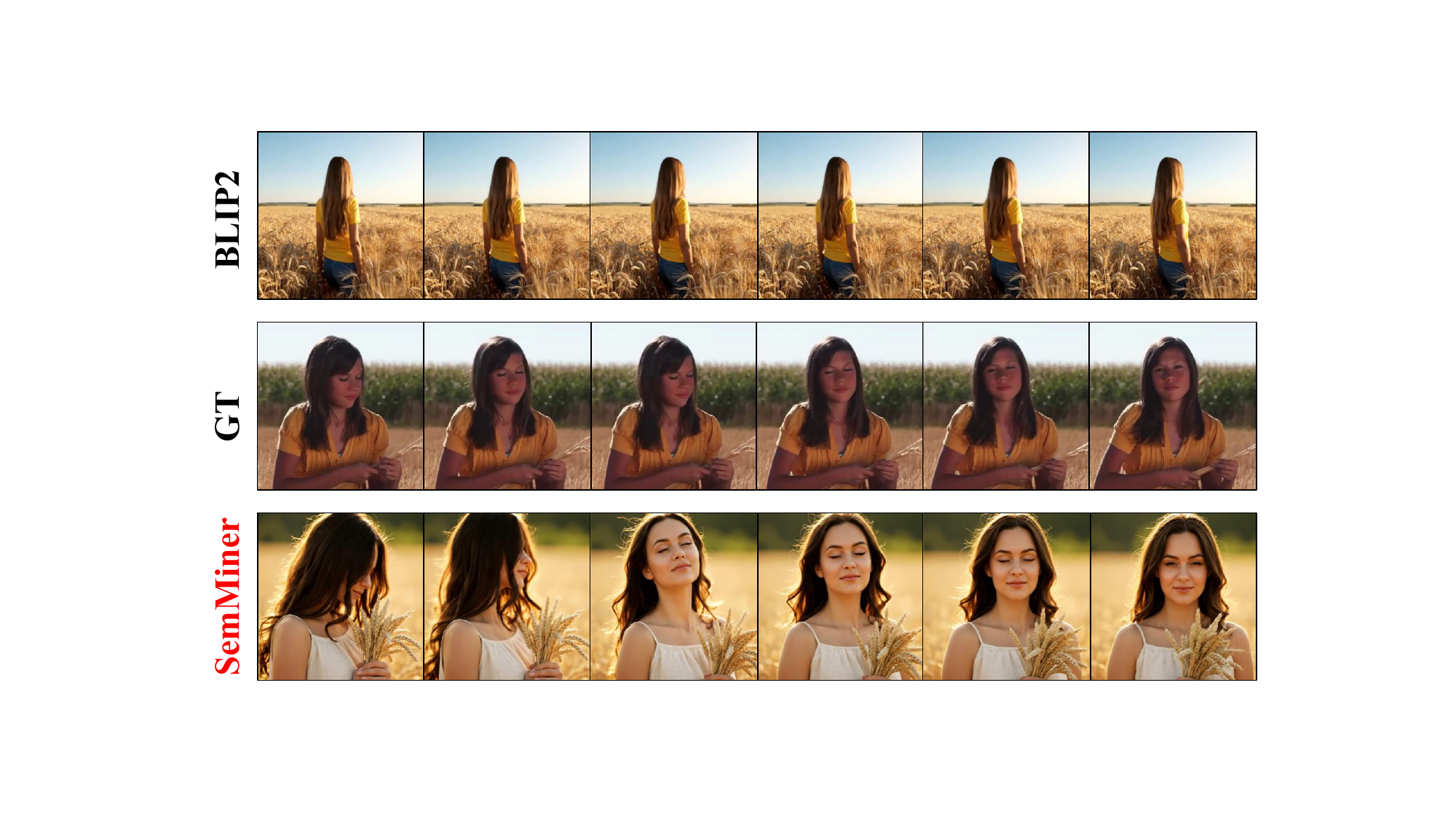}
    \vspace{-0.3cm}
    \caption{Video generated from BLIP2 caption(upper) and SemMiner(lower). The middle is the target video.}
    \label{fig:semprompt_video}
\end{figure*}
In this section, we randomly sample a video from the CC2017 dataset to qualitatively evaluate the descriptions generated by \Semprompt. As shown in Figure.\ref{fig:prompt} and Figure.\ref{fig:semprompt_video}, we first present the semantic caption produced by BLIP2\cite{xue2024xgenmmblip3familyopen}, a widely-used video description model in prior fMRI-based video reconstruction works\cite{gong2024neuroclips,lu2025animate}. It is evident that BLIP2 generates a coarse, static scene description and fails to capture the temporal dynamics of the video. In contrast, the multi-level descriptions generated by \Semprompt~—namely $C_{anchor}$, $C_{motion}$, and $C_{holi}$—are significantly more fine-grained and informative. The $C_{anchor}$ description accurately captures the visual content of the initial frame, identifying not only salient objects and scene types (\textit{e.g.}, ``a young woman standing in a wheat field'') but also detailing spatial layout and appearance features (\textit{e.g.}, ``wearing a short-sleeved blouse of a distinct mustard-yellow color and gently cradling a small bundle of golden-brown wheat stalks in her hands''). Similarly, $C_{motion}$ and $C_{holi}$ provide coherent and semantically rich accounts of the dynamic processes and holistic scene context, respectively, demonstrating the effectiveness of \Semprompt~in capturing multi-faceted video semantics.

Additionally, we quantitatively compare the average number of tokens in the descriptions generated by BLIP2 and \Semprompt, as shown in Table~\ref{tab:number}. It is evident that BLIP2 produces relatively short captions, with an average of only 10.6 tokens. In contrast, \Semprompt~ generates significantly richer descriptions, with averages of 32.8, 41.1, and 52.8 tokens for $C_{anchor}$, $C_{motion}$, and $C_{holi}$, respectively. These results highlight the ability of \Semprompt~to provide more detailed and semantically informative supervision signals, which in turn facilitate higher-quality neural decoding and video reconstruction.

 \begin{table}[htbp]
     \centering
     \renewcommand{\arraystretch}{1.1}
     \resizebox{0.48\textwidth}{!}{
         \begin{tabular}{l|cccc}
             \toprule
             caption type & BLIP2 & $C_{anchor}$ & $C_{motion}$ & $C_{holi}$ \\
             \midrule
              number    & 10.6 & 32.8 & 41.1 & 52.8 \\
             \bottomrule
         \end{tabular}
     }
     \caption{Average number of tokens in the descriptions generated by BLIP2 and \Semprompt~on CC2017 Datasets.}
     \label{tab:number}
 \end{table}

\subsection{Caption Similarity Analysis}
To quantitatively assess the semantic gap between the captions generated by \Semprompt, and to verify that each description indeed captures a distinct perspective of the video stimulus, we evaluate the semantic similarity among them using a set of standard language similarity metrics~\citep{papineni-etal-2002-bleu,hessel-etal-2021-clipscore}. As shown in Table~\ref{tab:Similarity_Analysis_Metrics}, the results reveal clear semantic divergence across the generated descriptions ($C_{anchor}$, $C_{motion}$, and $C_{holi}$), confirming that \Semprompt~effectively extracts complementary and perspective-specific semantic information from the input video.
 \begin{table}[htbp]
     \centering
     \renewcommand{\arraystretch}{1.1}
     \resizebox{0.48\textwidth}{!}{
         \begin{tabular}{l|cccccc}
             \toprule
             Pair & BLEU-1 & BLEU-4 & METEOR & ROUGE-1 & ROUGE-L & CLIP-S \\
             \midrule
              $C_{anchor}$ v.s. $C_{motion}$    & 0.171 & 0.022 & 0.186 & 0.275 & 0.207 & 0.655 \\
             $C_{anchor}$ v.s.  $C_{holi}$     & 0.197 & 0.028 & 0.222 & 0.304 & 0.225 & 0.665 \\
             $C_{motion}$ v.s.  $C_{holi}$  & 0.270 & 0.067 & 0.292 & 0.400 & 0.293 & 0.744 \\
             \bottomrule
         \end{tabular}
     }
     \caption{Semantic Similarity Experiments of different caption pairs. Lower values across all metrics indicate greater semantic dissimilarity.}
     \label{tab:Similarity_Analysis_Metrics}
 \end{table}
 
\subsection{The Impact of Different VLLM}
To evaluate the generalizability of \Semprompt, we integrate three different vision-language large models (VLLMs)—Video-LLaMA~\citep{zhang2023videollamainstructiontunedaudiovisuallanguage}, Qwen2.5-VL~\citep{bai2025qwen25vl}, and LLaVA-Video~\citep{zhang2025llavavideovideoinstructiontuning}—to generate semantic targets, using the same implementation details described in Appendix A.1.

The comparison results, presented in Table~\ref{tab:VLLM version}, show that while the optimal performance on individual metrics varies slightly across different VLLMs, all configurations achieve consistently strong and competitive reconstruction quality. The consistently high results confirm that the effectiveness of \Semprompt~ does not rely on any specific VLLM. This robustness is rooted in its core design — a hierarchical semantic prompting strategy that reliably guides diverse VLLMs to generate rich, structured, and multi-perspective video descriptions, thereby providing high-quality semantic supervision for fMRI-based video reconstruction.
 \begin{table*}[htbp]
	\centering
    \vspace{-2.5mm}
    \renewcommand{\arraystretch}{1.12}
	\resizebox{\textwidth}{!}{
		\begin{tabular}{l|ccccc|ccc|cc}
			\toprule
			\multirow{2}{*}{MLLM} & \multicolumn{5}{c|}{Semantic-Level} & \multicolumn{3}{c|}{Pixel-Level} & \multicolumn{2}{c}{ST-Level} \\
			& 2-way-I$\uparrow$ & 2-way-V$\uparrow$ & 50-way-I$\uparrow$ & 50-way-V$\uparrow$ & VIFI-score$\uparrow$ & SSIM$\uparrow$ & PSNR$\uparrow$ & Hue-pcc$\uparrow$ & CLIP $\uparrow$ & EPE $\downarrow$ \\
			\midrule
			VideoLlama     & \textbf{0.826} & 0.860 & \textbf{0.211} & 0.239 & 0.590 & \textbf{0.330} & \textbf{9.626} & 0.841 & \textbf{0.502} & 4.768 \\
			\ Qwen2.5-vl     & 0.821 & \textbf{0.863} & 0.203 & \textbf{0.242} & 0.589 & 0.299 & 9.03 & 0.858 & 0.496 & \textbf{4.40} \\
			\ Llava-Video       & 0.815 & 0.855 & 0.207 & 0.228 & \textbf{0.591} & 0.309 & 8.96 & \textbf{0.861} & 0.493 & 4.65 \\
			\bottomrule
		\end{tabular}
	}
    \vspace{-0.3cm}
    \caption{Ablation study on the CC2017 dataset investigating performance variations when using different VLLMs within the \Semprompt~ framework.}
    \label{tab:VLLM version}
\end{table*}
\begin{figure*}[htbp]
    \centering
    \includegraphics[width=0.98\textwidth, trim=80 60 20 60, clip]{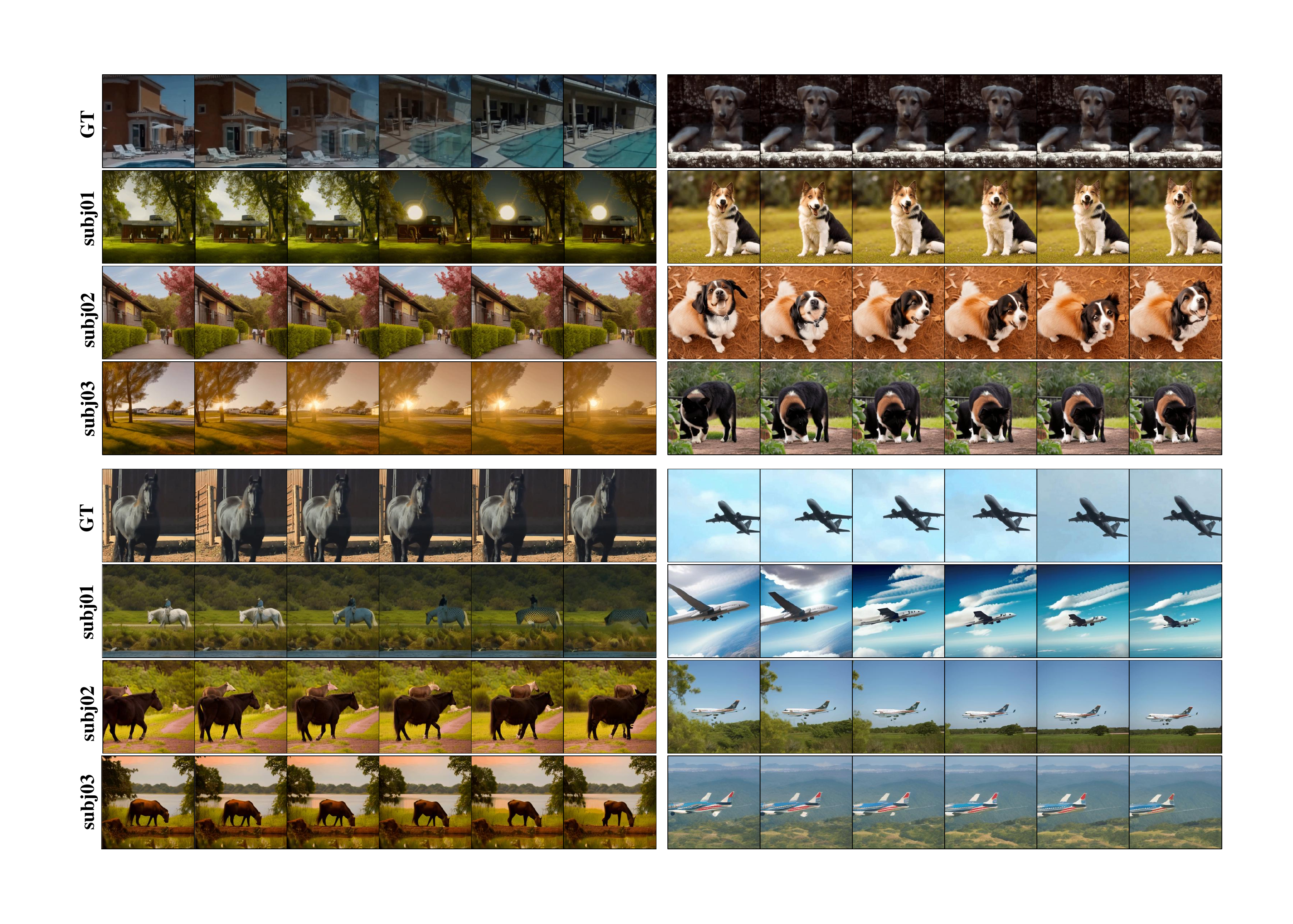}
    \vspace{-0.3cm}
    \caption{More reconstruction results on the CC2017 dataset from three subjects.}
    \label{fig:cross}
\end{figure*}
\begin{figure*}[htbp]
    \centering
    \includegraphics[width=0.98\textwidth, trim=80 60 20 60, clip]{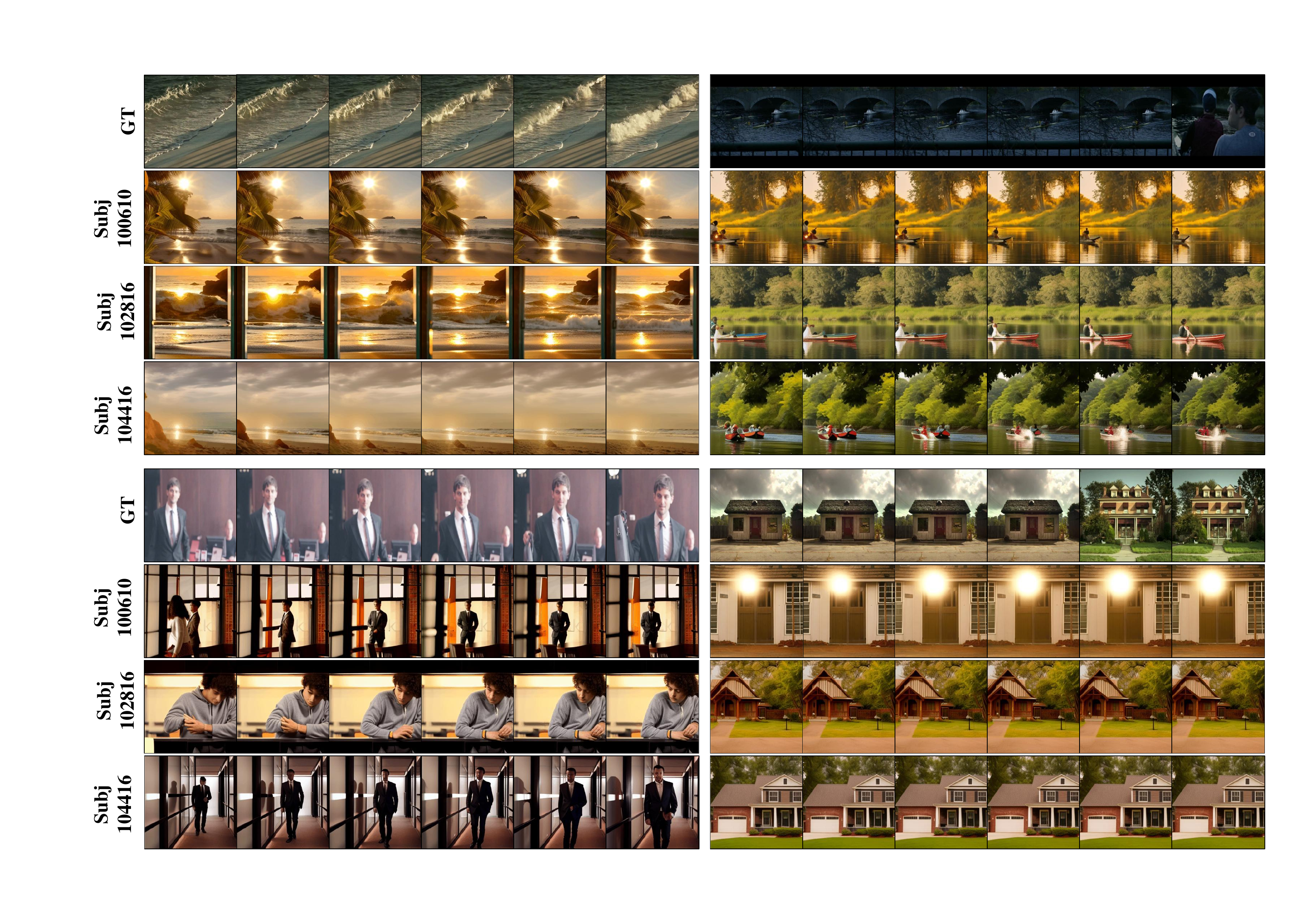}
    \vspace{-0.3cm}
    \caption{More reconstruction results on the HCP dataset from three subjects.}
    \label{fig:HCP}
\end{figure*}

\section{Evaluation Metrics}
We devised a multi-level evaluation protocol to comprehensively assess the performance of our video reconstruction model. This protocol systematically quantifies the fidelity of reconstructed videos to the original visual stimuli across three key dimensions: semantic, pixel, and spatiotemporal.

\subsection{Semantic-level}
Semantic-level metrics quantitatively assess the conceptual alignment between the reconstructed and original videos. We employ two methods for this purpose: the N-way Top-1 accuracy test and the VIFI-Score.
\textbf{N-way Top-1 Accuracy} measures the precision of semantic content reconstruction using a dual strategy: (1) Dynamic Content.  To evaluate the semantic fidelity of dynamic events, a VideoMAE~\cite{Tong2022VideoMAE} model pretrained on the Kinetics-400 dataset classifies both the original and reconstructed videos, after which the consistency of their predicted class labels is compared. 
(2) Static Content. To assess the accuracy of static elements, a ViT-For-Image-Classification model~\cite{dosovitskiy2021image} pretrained on ImageNet performs frame-by-frame classification.

\textbf{VIFI-Score}:This metric is calculated using a VIFICLIP~\citep{rasheed2023fine} model that has been fine-tuned on video datasets. The model extracts high-level feature embeddings from both the original and reconstructed videos. The final VIFI-Score is the cosine similarity between these two embedding vectors, where a higher value indicates greater semantic congruence.
\subsection{Pixel-level}
Pixel-level metrics quantitatively assess low-level visual fidelity by directly comparing the reconstructed video to the original visual stimulus. These metrics provide a foundational measure of reconstruction quality by focusing on the accuracy of pixel values and the preservation of fundamental image characteristics.

\textbf{Peak Signal-to-Noise Ratio (PSNR)}: A widely adopted metric for quantifying reconstruction error, PSNR is derived from the Mean Squared Error (MSE) between the pixel values of the original and reconstructed images and measures distortion on a logarithmic scale.

\textbf{Structural Similarity Index (SSIM)}: This metric evaluates the preservation of structural information between frames. Unlike purely pixel-wise error metrics, SSIM is designed to model perceptual similarity by jointly considering luminance, contrast, and structure.

\textbf{Hue Similarity}: This metric specifically assesses color reproduction accuracy by computing the cosine similarity between the hue channel vectors of the original and reconstructed frames. The hue component is isolated because it represents pure color, independent of saturation and brightness. A high score therefore indicates that the model has successfully reproduced the chromatic ambiance of the original scene, a critical factor for the overall visual experience.

\subsection{Spatiotemporal (ST)-level}
Spatiotemporal (ST) metrics are designed to evaluate the dynamic aspects of a reconstructed video, focusing on its temporal coherence and motion representation accuracy. These metrics are crucial for assessing a model's ability to generate realistic and continuous sequences of events.

\textbf{CLIP-pcc~\cite{wu2023tune}}: This metric assesses the temporal smoothness and semantic continuity of a video sequence. It is calculated by using a CLIP (ViT-B/32) model to extract feature embeddings for each frame and then averaging the cosine similarity across all adjacent frame pairs.

\textbf{Endpoint Error (EPE~\citep{Barron1994Performance})}: This metric provides a precise quantification of motion reconstruction fidelity. It is defined as the mean Euclidean distance between the optical flow vectors of the reconstructed and original videos. By directly comparing these predicted motion fields, EPE serves as a critical indicator of the model's capability to accurately reproduce the trajectories and velocities of objects within the scene.

\section{Implementation Details}

\subsection{Hyperparameter Settings}
For the Semantic Alignment Decoder, we utilized the AdamW~\cite{Loshchilov2017bsp} optimizer. The training was conducted for 100 epochs with a batch size of 96. A cosine annealing schedule was employed to manage the learning rate, with a maximum value set to 1e-4.  $\lambda_{\text{refine}}$ and $\lambda_{\text{SoftCLIP}}$ are set to 0.5 and 0.1, respectively. Furthermore, to mitigate the risk of overfitting arising from the scarcity of fMRI-video data pairs, we incorporated the Mixco data augmentation technique. 
 \begin{equation}
     x_{mix_{i,j}} = \alpha x'_i + (1 - \alpha) x'_j,
 \end{equation}
 where $\alpha \sim \text{Beta}(0.15, 0.15)$ is a mixing coefficient sampled from the Beta distribution, and $x'_i$ is the $i$-th latent representation to the $x'$-encoded batch.

For the Motion Adaptation Decoder, the AdamW optimizer was also employed, and the model was trained for 100 epochs with a batch size of 20. The learning rate was governed by a OneCycleLR~\cite{smith2018superconvergencefasttrainingneural} schedule, with the maximum learning rate configured to 3e-4. Notably, $\lambda_{\text{spat}}$ and $\lambda_{\text{temp}}$, were not pre-defined hyperparameters. Instead, they were implemented as learnable parameters that are automatically optimized during the training process. 
\subsubsection{Model Architecture}
For Semantic Alignment Decoder, we use a linear layer to implement the subject-specific mapper $f_{\text{SAD}}^{\theta_m}$, which maps voxels $x \in \mathbb R^{D_m}$ to the shared space $\mathbb R^{4096}$.
$f_{\text{SAD}}^{\textit{MLP}}$ is implemented as a four-layer perceptual machine
with output channels specified as  $\{4096, 4096, 4096, 77 \times 768\}$, using the GELU nonlinear activation function.
The ($f_{\text{SAD}}^{\textit{Refine}}$) module uses a Transformer architecture with 4 layers, each containing 8 attention heads of dimension 64. 

For the Motion Adaptation Decoder, $f_{\text{MAD}}^{\textit{proj}}$ is composed of a linear layer and a four-layer Multi-Layer Perceptron with output channels specified as  $\{256, 6\times4096, 6\times4096, 4096, 6 \times (64*7*7)\}$.The fusion attention module is a lightweight decoder with a UNet-like~\cite{ronneberger2015unetconvolutionalnetworksbiomedical} architecture, consisting of one UNetMidBlock2D and three cascaded AttnUpDecoderBlock2D upsampling blocks. The architecture takes a 64-channel, 7x7 feature map as input and progressively upsamples it to a 4-channel, 28x28 latent representation.
\begin{figure*}[htbp]
    \centering
    \includegraphics[width=0.98\textwidth, trim=0 0 0 0, clip]{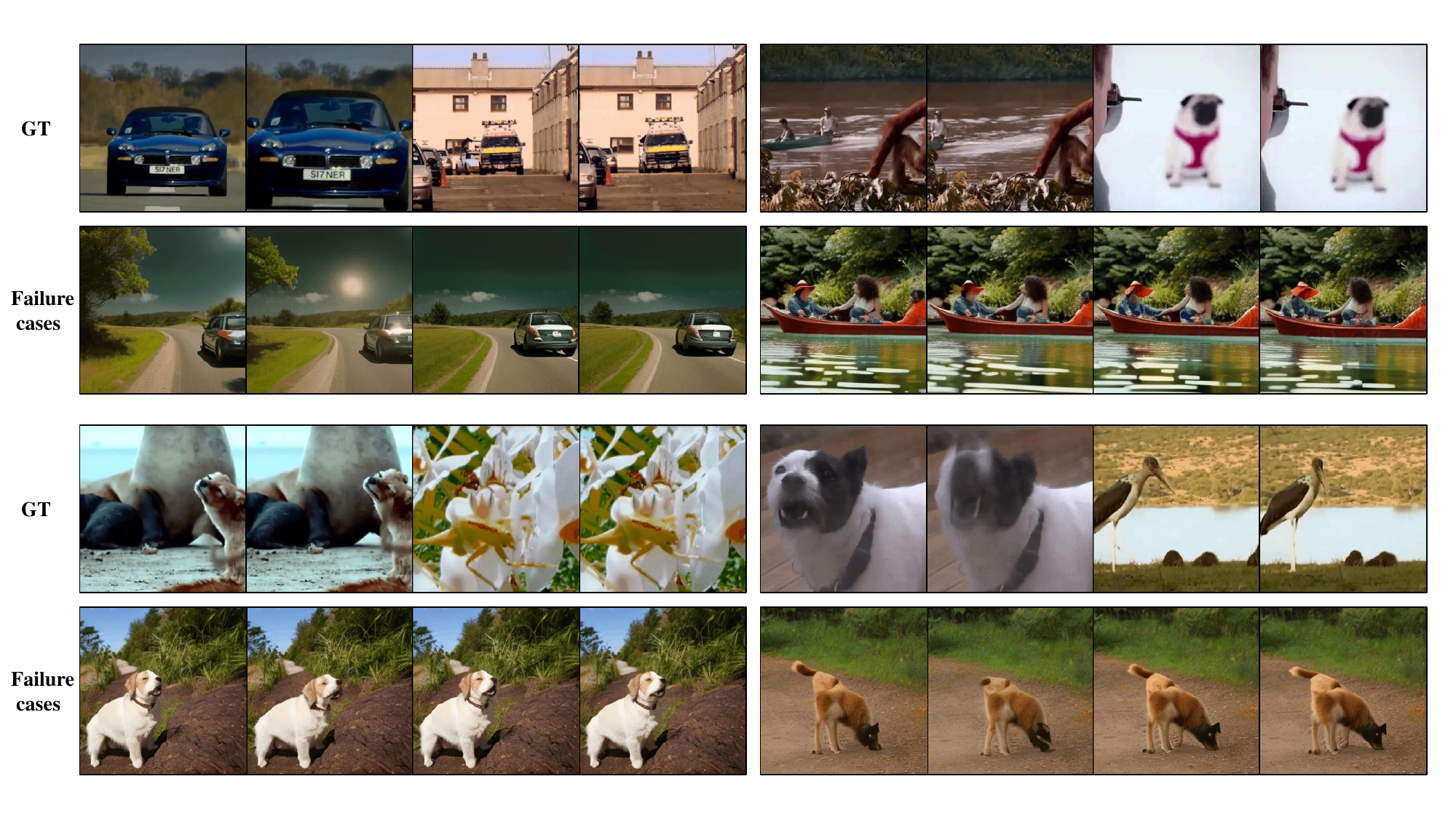}
    \vspace{-0.6cm}
    \caption{Reconstruction failure cases.}
    \label{fig:failure}
\end{figure*}
\section{More Results}
We present further examples of video reconstruction across multiple subjects from both the CC2017~\cite{Wen2017Neural} and HCP~\cite{marcus2011informatics} datasets, as illustrated in Figure.\ref{fig:cross} and Figure.\ref{fig:HCP}.
\begin{figure*}[t] 
    \centering
    
    \includegraphics[width=0.98\textwidth, trim=0 180 0 180, clip]{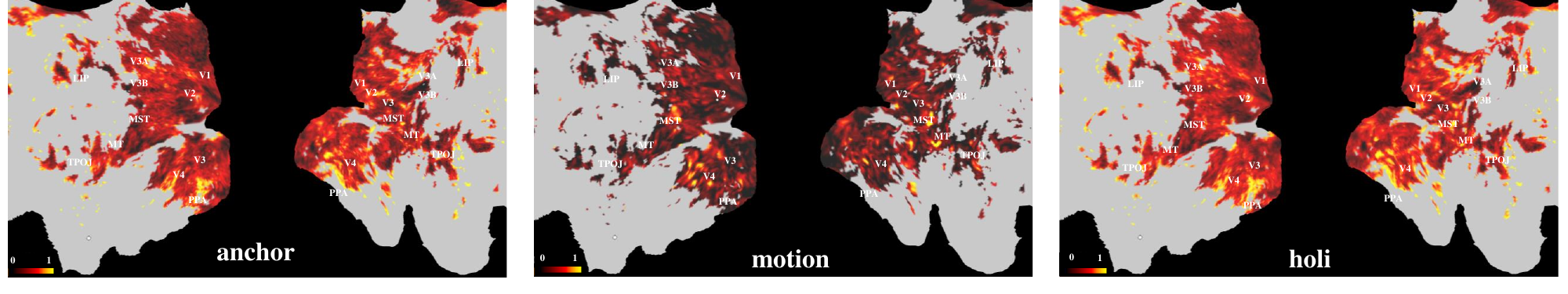}
    \vspace{-0.3cm}
    \caption{Importance visualization of different ROIs for Subject 2, based on the fitted weights from the first layer of the SFD. Weights from each module are averaged and normalized to the $[0, 1]$ range for comparison.}
    \label{fig:cortex_2}
    
    \vspace{10pt} 
    
    \includegraphics[width=0.98\textwidth, trim=0 180 0 180, clip]{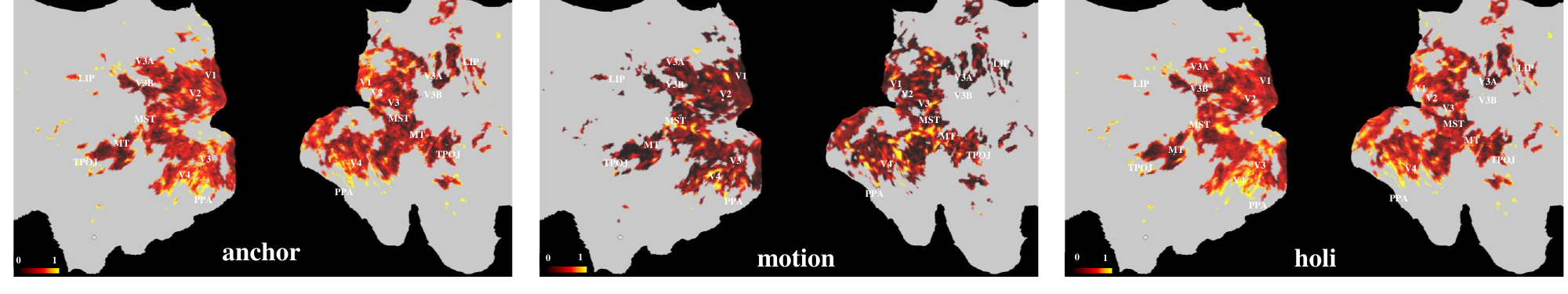}
    \vspace{-0.3cm}
    \caption{Importance visualization of different ROIs for Subject 3, based on the fitted weights from the first layer of the SFD. Weights from each module are averaged and normalized to the $[0, 1]$ range for comparison.}
    \label{fig:cortex_3}
    
\end{figure*}
\subsection{Neural Interpretability}
To further validate the neural interpretability of the proposed framework, we generated voxel-wise importance maps of the cerebral cortex for Subjects 2 and 3 from the CC2017 dataset, as depicted in Figure.\ref{fig:cortex_2} and Figure.\ref{fig:cortex_3}. The experimental results for these subjects reveal voxel importance distribution patterns that are highly similar to those observed for Subject 1: The anchor component's activations were predominantly concentrated in visual cortical areas. The motion component exhibited stronger activations in brain regions specialized for motion processing. The holistic component displayed a balanced activation pattern across both of these functional regions. This hierarchical semantic mapping enables \Semvideo~to generate more perceptually faithful reconstructions and similar activation distributions across subjects demonstrate stable feature capture for both static semantics and dynamic behaviors.
\subsection{Failure cases}
To facilitate a comprehensive and objective assessment of $SemVideo$, we analyze several of its reconstruction failure cases in Figure.\ref{fig:failure}. A primary cause of these failures stems from the inherent data acquisition paradigm. Specifically, the experimental data is created by evenly segmenting longer videos, which introduces abrupt content transitions at the boundaries of the resulting clips. Our model, $SemVideo$, struggles to replicate these sudden shifts in the reconstructed outputs.


\end{document}